\documentclass[]{fairmeta}

\usepackage{amsmath,amssymb}
\usepackage{booktabs}
\usepackage{placeins}
\usepackage{tabularx}


\title{On BatchNorm Forward Modes in \\ Value-Based Reinforcement Learning}

\author[1,2,3]{Daniel Palenicek}
\author[1]{Mikael Henaff}
\author[1]{Scott Fujimoto}
\author[1]{Koustuv Sinha}

\affiliation[1]{FAIR at Meta}
\affiliation[2]{Technical University of Darmstadt}
\affiliation[3]{hessian.AI}

\abstract{
Batch normalization (BN) substantially improves sample efficiency in continuous-control actor-critic methods such as CrossQ, yet recent studies report performance degradation in discrete-action value learning on Atari.
These failures are surprising because discrete Q-networks lack the action-input distribution mismatch identified by CrossQ.
We show for target-based C51 and target-free PQN that the simple choice between running and batch statistics at specific forward passes can reverse this degradation.
In C51, switching the BN bootstrap forward to batch-statistic mode significantly improves performance over unnormalized and LayerNorm baselines and scales stably with update-to-data ratios up to 12.
In PQN, using batch-statistics for both action selection and bootstrapping recovers performance from the failing running-statistic configuration.
Across 26 Atari games at 400M frames, this configuration achieves a higher final aggregate score than PQN with LayerNorm.
Our results show that carefully configured BN can substantially improve discrete-action value learning, and that its forward protocols are an essential part of the algorithm specification.

}

\correspondence{Daniel Palenicek at \email{palenicek@robot-learning.de}}

\begin{document}

\maketitle

\section{Introduction}
\label{sec:introduction}

Before CrossQ~\citep{bhatt2024crossq}, batch normalization (BN) was widely regarded as ineffective or unstable in deep reinforcement learning (RL), with prior works reporting degradation in learning performance~\citep{salimans2016weightnorm,hiraoka2022dropout}.
CrossQ overturned this empirical picture in continuous-control actor--critic learning, showing that a carefully constructed BN forward pass can substantially improve sample efficiency. It evaluates replay and current-policy state--action pairs in one joined critic forward pass so that their different action distributions are normalized jointly.
Subsequent methods successfully retain BN-based architectures while reintroducing target critics, adding fixed weight-norm projections, incorporating categorical critics, scaling to high-dimensional robotic control, leveraging prior expert-data and parallel simulators~\citep{palenicek2025crossqwn,palenicek2026xqc,palenicek2026xqcfd,kim2026flashsac}.
Together, these results show that BN can be highly effective in continuous-control RL when its interaction with the critic update is specified carefully.

The current discrete-action value-based literature presents a sharply different picture. 
\citet{salimans2016weightnorm} report that minibatch-statistic noise destabilized DQN and that they could not make BN work without impractically large minibatch sizes.
Later studies report BN degradation in DQN on MinAtar~\citep{gogianu2021spectral},
in PQN on Atari~\citep{gallici2025simplifying},
and in a five-game Atari comparison~\citep{vincent2025bridging}.

These conflicting results are particularly surprising since the action distribution failure mode identified by CrossQ is entirely absent in discrete-action value learning.
A continuous-action actor-critic critic-network receives both states and actions, $Q(s,a)$, and therefore processes replay actions and current-policy actions drawn from different populations.
A discrete-action critic-network instead receives only a state and emits a vector of action-values, $Q(s,\cdot)$.
Therefore, the action-input mismatch cannot explain the reported failures.

We show, that a different mismatch can arise between the stored BN running statistics and the activations encountered by a particular forward pass.
When state distributions evolve during training, the stored running-statistics can become poorly matched to the target network’s current activations.
More generally, prediction, bootstrapping, and action selection can each evaluate the network on different input populations. This motivates our central question: \emph{can the choice between running and batch statistics at these different forward passes determine whether BN helps or harms discrete-action value learning?}

We investigate this question in two complementary settings.
C51 uses replay and periodically hard-copied target parameters, allowing us to study the bootstrap forward mode while retaining a conventional target-network architecture (\Cref{sec:c51-results}). 
Our primary comparison changes the target-network BN forward mode from stored running statistics to statistics computed on the current next-state minibatch.
We then ablate how quickly stored target moments track the bootstrap population.
This sweep tests whether the performance gap between running statistics and batch statistics is accompanied by systematic sensitivity to the responsiveness of running statistics.
Having established the effect of the bootstrap protocol, we examine its effectiveness for replay reuse.
We fix training-mode bootstrapping and demonstrate how different C51 configurations respond as the update-to-data ratio (UTD) increases. BN based C51 variants significantly outperform the baselines, especially with increasing UTD.
PQN provides a complementary setting with neither replay nor a separate target network~(\Cref{sec:pqn-results}).
It allows us to test whether sensitivity to BN forward protocols persists when those two features
of C51 are absent.
We demonstrate that the same switch to batch statistics recovers performance and the BN-based PQN variant outperforms vanilla PQN on the full 400 million frame budget.

Our experiments establish three main findings:
\begin{enumerate}
    \item \textbf{Bootstrap statistics can reverse BN degradation in C51.}
    Across 26 Atari games, replacing stored target moments with current-minibatch statistics produces large performance improvements and changes BN from underperforming vanilla C51 to significantly exceeding its performance. A tracking sweep further shows that performance depends on the responsiveness of stored target moments.

    \item \textbf{Batch-statistic bootstrapping supports effective learning with increased replay reuse.} Across UTDs $\in\{0.25,1,2,4,8,12\}$, BN configurations continue to benefit from additional updates beyond the range where unnormalized C51 and LayerNorm peak. The projected-BN configuration remains effective through UTD 12.

    \item \textbf{Forward-protocol sensitivity persists in target-free PQN.} Using batch statistics for action selection and bootstrap-target construction recovers performance from the failing running-statistic configuration originally reported in \citet{gallici2025simplifying}. Across 26 games at 400M frames, the selected configuration achieves comparable performance over training and a higher final score than LayerNorm.

\end{enumerate}
Together, these findings show that carefully configured BN can substantially improve discrete-action value learning.
They also explain why specifying that an agent ``uses BN'' is insufficient: each prediction, bootstrap, acting, and evaluation forward must define its parameter source, input batch, normalization statistics, and treatment of running-state updates.
These choices need to be carefully considered as they determine the function evaluated by the agent and can significantly impact performance.

\section{Value Learning and BatchNorm Semantics}
\label{sec:study-design}

We consider a Markov decision process with finite action set $\mathcal{A}$ and discount $\gamma\in[0,1)$~\citep{sutton2018reinforcement}. Deep Q-learning represents the action-value function with a network $Q_\theta$~\citep{mnih2015dqn}. For a transition $(s,a,r,s',d)$, a one-step TD target has the form
\begin{equation}
\textstyle y= r+\gamma(1-d)\max_{a'}Q_{\bar\theta}(s',a'),
\label{eq:dqn-target}
\end{equation}
where $\bar\theta$ may be a separate target network.
C51 implements the target distributionally with a categorical Bellman projection and hard-copied target parameters~\citep{bellemare2017c51}. PQN instead constructs multi-step $Q(\lambda)$ targets from parallel rollouts without replay or target parameters~\citep{gallici2025simplifying}.

For a BN layer $\ell$, the relevant moments are properties of a network--population pair,
\begin{equation}
\textstyle M_\ell(\theta,\mathcal{D})=
\left(
\mathbb{E}_{x\sim\mathcal{D}}[h_\ell(x;\theta)],
\operatorname{Var}_{x\sim\mathcal{D}}[h_\ell(x;\theta)]
\right),
\label{eq:network-population-moments}
\end{equation}
not of $\theta$ alone~\citep{ioffe2015batch}. A batch-statistic forward estimates these moments from its current input batch and uses them in the current output. A running-statistic forward instead reads an exponential average of moments produced by earlier network--batch pairs. This historical mixture is usually meaningful when supervised training repeatedly samples a stationary dataset and is followed by a separate deployment phase. In RL, network parameters and input populations co-evolve. Policy visitation changes, replay turns over, and prediction, bootstrap, and acting forwards consume different populations. Running statistics therefore chase a moving network--population target, as they can in continual learning~\citep{pham2022continual}.

We characterize each forward by its network parameters, input population, statistic source, and whether it updates stored BN statistics.
Framework ``training'' and ``evaluation'' modes typically couple the choice of statistics with updates to these buffers.
Copying online parameters and BN buffers to a target network does not guarantee that the copied moments match the activations encountered by subsequent target forwards.

\section{Bootstrap Forward Modes and Replay Scaling in C51}
\label{sec:c51-results}

\begin{figure*}[t!]
    \centering
    \includegraphics[width=\textwidth]{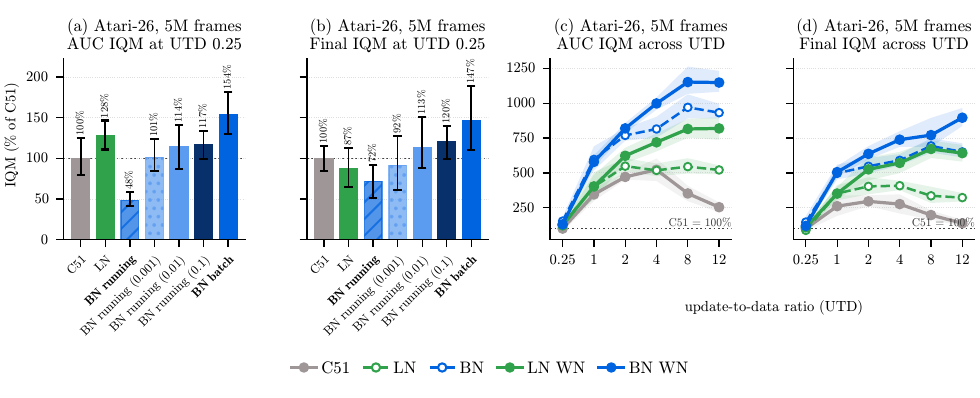}
    \caption{\textbf{Bootstrap forward modes and replay scaling in C51.}
    All panels report performance as a percentage of unnormalized C51 at UTD $0.25$, with C51 fixed to 100\%. Panels (a,b) report AUC and final IQM at UTD $0.25$. \emph{BN running} uses evaluation-mode bootstrapping and reads stored target moments. \emph{BN batch} uses training-mode bootstrapping and statistics from the current next-state minibatch. The three \emph{BN running ($\alpha$)} conditions update stored target moments from that minibatch with PyTorch update weight $\alpha\in\{0.001,0.01,0.1\}$ before constructing the target in a separate evaluation-mode forward. Panels (c,d) report AUC and final IQM across UTD for C51, LayerNorm (LN), BN, and their fixed-norm projection variants (WN). C51, LN, and BN use a constant learning rate, whereas LN WN and BN WN use linear decay. Every displayed bar or arm--UTD point contains 26 games and three seeds per game, each trained for 5M frames. Bars and curves show IQM; error bars and bands are marginal 95\% stratified bootstrap confidence intervals across 26 Atari games and 3 seeds each.}
    \label{fig:c51-main}
\end{figure*}

Our C51 study follows the two stages shown in \Cref{fig:c51-main} from left to right.
First, we ask whether the bootstrap-forward mode changes the negative BN result. We compare evaluation mode, which reads stored running moments, with training mode, which uses statistics from the current bootstrap minibatch. 
Additionally, we use a tracking-rate sweep to interpolate between them.
Second, having fixed the direct training-mode bootstrap, we ask how complete C51 configurations respond to increased replay reuse.

These experiments are based on the CleanRL C51 implementation~\citep{huang2022cleanrl,bellemare2017c51}.
We run every agent for 5M frames on all 26 games with three seeds per game.
We aggregate human-normalized scores with the interquartile mean (IQM)~\citep{agarwal2021rliable}.
Area under the learning curve (AUC) averages each run's equally spaced evaluation grid, and the final statistic uses the evaluation at 5M frames.
Confidence intervals use 5,000 stratified bootstrap resamples. Figure values are percentages of the unnormalized C51 point estimate at UTD $0.25$, not percentage improvements. 

\subsection{Evaluation-mode versus training-mode bootstrapping}
\label{sec:c51-bootstrap}

When augmenting C51 with BN using running statistics, we see a significant drop in performance (\Cref{fig:c51-main} (a,b)) to only 48\% AUC and 72\% final performance.
This reproduces the reported failures of BN in value-based RL.
Simply switching the target BN layer to training mode, using the batch statistics, performance increases dramatically to 154\% AUC of C51 and 147\% Final, respectively.
All other hyperparameters remain at standard CleanRL settings.
In summary, evaluation-mode bootstrapping reproduces the negative aggregate BN result, whereas training-mode bootstrapping significantly improves performance over C51.
The training-mode BN even outperforms a C51 LN baseline.

\subsection{Tracking interpolation between the endpoint modes}

To further analyze the gap between fixed running statistics and batch statistics, 
we add an additional synthetic ablation to the failing BN running experiment (\Cref{fig:c51-main} (a,b)).
In this, we add an additional training-mode forward pass right before the running statistics forward pass.
Before each bootstrap, a discarded training-mode pass over the next-state minibatch updates the target buffers with PyTorch update weight $\alpha\in\{0.001,0.01,0.1\}$; a second, evaluation-mode pass then constructs the target.
Larger $\alpha$ makes the stored moments more responsive to the current batch population. This experiment essentially interpolates between relying on historical running moments and relying on the current batch. 

Interestingly, performance monotonically increases the more \emph{online} the running statistics get to the batch statistics. Using the actual batch statistics yields the strongest results and results in the most straight forward implementation.
Making evaluation-mode bootstrapping work requires maintaining running statistics, choosing a tracking coefficient, and performing an additional update pass before constructing the target. Direct training-mode bootstrapping uses a single target call, requires no tracking hyperparameter, and makes the target output independent of historical running moments. 
Because direct training mode is both the simplest implementation and the strongest displayed AUC endpoint, we fix it for the replay-scaling study.

\subsection{Scaling with replay reuse}
\label{sec:c51-utd}

With training-mode bootstrapping fixed, we sweep over UTD $\in\{0.25,1,2,4,8,12\}$ for unnormalized C51, LN, BN, and LN and BN configurations with WN fixed-norm projection~\citep{lyle2024normalization,palenicek2025crossqwn}.
Every arm--UTD cell contains complete evaluation histories through 5M frames for all 26 games and three seeds. The WN variants in \Cref{fig:c51-main} (c,d) use linear learning-rate decay, whereas the three unprojected configurations use a constant learning rate.

Increasing replay reuse produces markedly different results across these configurations. Unnormalized C51 and LayerNorm improve through UTD 2, plateau through UTD 4, and then lose aggregate performance as the number of updates rises. Plain BN continues to improve in AUC through UTD 8 and remains above both unprojected controls at UTD 12 despite declining at the final point. Projected LayerNorm also benefits substantially from replay reuse, while projected BN shows the strongest high-UTD response: its AUC rises from about $1.3\times$ the UTD-$0.25$ C51 reference to $11.5\times$ at UTD 8 and remains at that level at UTD 12. At UTD 8, projected BN has a higher three-seed per-game mean than matched unnormalized C51 on 25 of 26 games by AUC and all 26 games by final score.

Taken together, the C51 results establish a sequence. Evaluation-mode bootstrapping reproduces severe BN degradation; increasing the responsiveness of stored moments interpolates toward the stronger batch-statistic result; and the simplest endpoint, a direct training-mode bootstrap, supports effective complete configurations as replay reuse increases. We next ask whether role-specific sensitivity persists when replay and separate target parameters are removed.

\section{Forward Modes in Target-Free PQN}
\label{sec:pqn-results}

PQN~\citep{gallici2025simplifying} provides a complementary test because it has neither a replay buffer nor a separate target network and the authors report BN to degrade performance in their experiments.
The same parameters are reused for prediction, $Q(\lambda)$ target construction, and acting, but these calls still consume different input populations. In particular, vectorized acting introduces a setting that is absent from the C51 intervention. We therefore ask which acting and bootstrap modes reproduce the reported BN failure, then test a simple selected protocol in a separate 400M-frame Atari-26 experiment.
We use the official implementation of \citet{gallici2025simplifying}, which includes their BN ablation code, for a fair analysis.

\begin{figure}[t!]
    \centering
    \includegraphics[width=\textwidth]{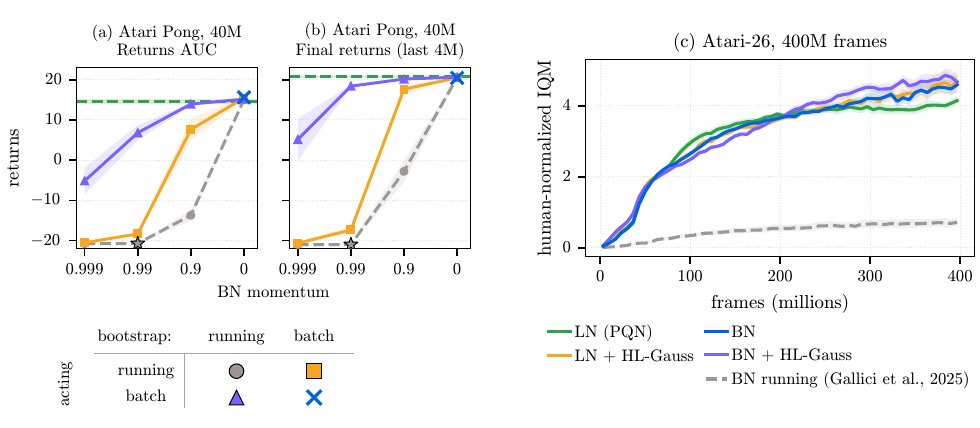}
    \caption{\textbf{Forward-mode sensitivity in target-free PQN.}
    Panels (a,b) report AUC and mean return over the last 4M frames of a 40M-frame Pong ablation. Gray uses running statistics for acting and bootstrapping, orange changes only bootstrapping to batch statistics, purple changes only acting to batch statistics, and blue uses batch statistics for both. The horizontal green line is LayerNorm; the gray star marks the released running/running configuration at Flax momentum $0.99$. Lines and points show means, and shaded regions show standard errors across five seeds each.
    Panel (c) full IQM training curves over 400M frames on Atari-26 aggregated across ten seed.
    Intervals are 95\% stratified bootstrap confidence intervals within-game seed-bootstrap intervals.
    The dashed gray curve is the released running-statistic BN configuration from \citet{gallici2025simplifying};
    green and blue show scalar-regression LayerNorm and batch-statistic BN, and orange and purple show their respective HL-Gauss variants.}
    \label{fig:pqn-main}
\end{figure}

\subsection{Localizing acting and bootstrap sensitivity on Atari Pong}

In \Cref{fig:pqn-main} (a,b), we replace PQN's LayerNorm layers with BN and cross the statistic sources used for acting and bootstrapping; loss predictions always use batch statistics. Our 40M-frame Pong diagnostic compares running statistics for both roles, batch-statistic bootstrapping only, batch-statistic acting only, and batch statistics for both.
Wherever a forward reads stored moments, we vary Flax momentum over $\{0,0.9,0.99,0.999\}$ in comparison (note that Flax momentum convention is flipped compared to PyTorch).
AUC averages 30 equal frame bins, the final statistic averages the last 4M frames. We run five seeds per configuration.

The running/running configuration first reproduces the failure \citet{gallici2025simplifying} report (\Cref{fig:pqn-main} (a,b)). At the released momentum of $0.99$, marked by the gray star, performance remains at random-policy performance and momentum $0.999$ behaves similarly.
Reducing momentum to $0.9$ is insufficient, whereas momentum zero recovers to approximately LayerNorm performance. 
Thus, the negative result is not invariant to adding BN: it depends strongly on the forward protocol and the responsiveness of the stored moments.

Changing the roles separately localizes this sensitivity. At momentum $0.9$, changing only bootstrapping to batch statistics raises AUC from $-13.70$ to $7.63$ and final return from $-2.77$ to $17.65$, but the configuration remains near the floor at momenta $0.99$ and $0.999$. Changing only acting to batch statistics is more robust across this Pong sweep: it approximately matches LayerNorm at momenta $0$ and $0.9$ (AUC $15.17$ and $13.94$; final return $20.62$ and $20.17$) and retains a positive final return at momentum $0.99$ ($18.40$). These point estimates identify acting as the more sensitive remaining role in this diagnostic, while also showing that changing bootstrapping alone can provide a substantial partial recovery.

Using batch statistics for both acting and bootstrapping reaches an AUC of $15.62$ and a final return of $20.45$. Because prediction also uses batch statistics, no forward pass in this configuration depends on running statistics, and the configured BN momentum is therefore irrelevant to its outputs.
This all-batch protocol is both the simplest tested configuration and the strongest Pong AUC point estimate, so we select it for the full-budget experiment.

\subsection{PQN BN on Atari-26 at 400M frames}
\label{sec:pqn-evaluation}

We next evaluate the selected all-batch protocol on all 26 Atari games using ten random seeds each. 
The primary comparison uses PQN's standard scalar-regression objective and contrasts batch-statistic BN with LayerNorm.
For comparison, we re-run \citet{gallici2025simplifying}'s released running-statistic BN configuration to anchor the original negative result.
We also add an additional HL-Gauss loss variant, where we replace the scalar regression loss with a Gaussian-smoothed categorical target and cross-entropy loss~\citep{imani2018improving,farebrother2024stop}.

The running-statistic BN configuration remains poor throughout the full horizon (\Cref{fig:pqn-main}c) as previously reported by \citet{gallici2025simplifying}.
Batch-statistic BN and LayerNorm attain final IQMs of $4.59$ and $4.14$, respectively.
Under the tested online protocol, batch-statistic BN is therefore competitive with LayerNorm over the course of training and has a significantly higher final return.
The matched HL-Gauss comparison provides a secondary objective check, with a slightly closer gap. While BN has a slight edge during training, both BN and LN final IQMs are $\sim 4.67$ for HL-Gauss.
Therefore, we demonstrate, that BN is competitive in PQN, if applied correctly.

\subsection{What PQN BN establishes}

The PQN experiments extend the forward-mode result beyond hard-target replay learning.
The previously released running/running configuration reproduces severe BN degradation. 
Our Pong diagnostic shows that acting and bootstrap calls contribute differently to that degradation, and the selected all-batch protocol is competitive with LayerNorm over 400M frames.
In particular, PQN shows that role-specific forward semantics remain consequential even when prediction, bootstrapping, and acting reuse one parameter set.
The evidence therefore supports BN as a viable target-free PQN configuration under the tested protocol, not general BN superiority over LayerNorm.

\begin{figure}[t!]
    \centering
    \includegraphics[width=\textwidth]{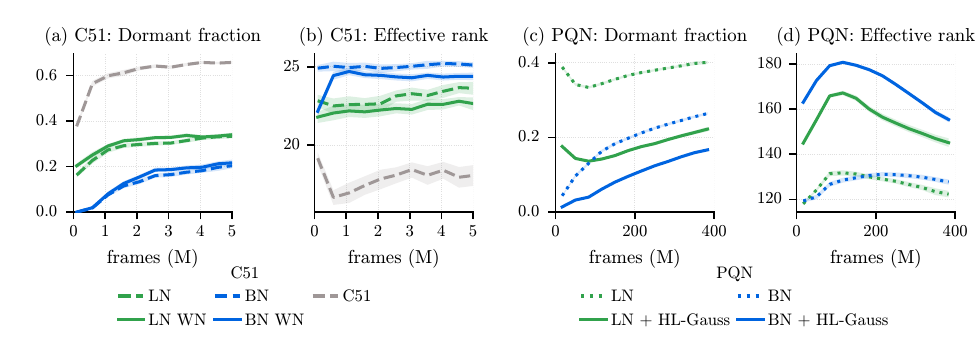}
    \caption{\textbf{Representation diagnostics in hard-target C51 and target-free PQN.}
    For each of the five C51 arms, panels (a,b) use all 78 matched game--seed identities at UTD 4.
    C51, LayerNorm, and BN use a constant learning rate and the two WN-projection arms (LN WN, BN WN) use a linear-decay learning rate, matching \Cref{fig:c51-main}.
    Panels (c,d) average PQN's ten per-game seeds across 26 games. 
    Both algorithms compute effective rank from centered post-ReLU 512-dimensional penultimate features. 
    Confidence intervals are pointwise 95\% stratified bootstrap confidence intervals obtained by resampling seed identities within each fixed game and recomputing the IQM.
    Color identifies the normalizer; in the C51 panels line style denotes the fixed-norm projection (WN), and in the PQN panels line style compares scalar regression and HL-Gauss for both normalizers. 
    }
    \label{fig:combined-plasticity}
\end{figure}

\section{Plasticity Diagnostics}
\label{sec:representation}

\citet{sokar2023dormant} introduced the dormant-neuron score in deep RL, and prior value-learning work connected additional gradient updates to feature-rank collapse under bootstrapping~\citep{kumar2021implicit}. Subsequent work used dormant-unit counts and feature-rank summaries to analyze loss of plasticity under nonstationarity and normalization~\citep{lyle2024normalization}. CrossQ+WN subsequently tracked dead neurons together with parameter, gradient, and effective-learning-rate statistics when studying replay scaling, while XQC analyzed Hessian spectra and related conditioning statistics~\citep{palenicek2025crossqwn,palenicek2026xqc}. We include the same class of diagnostics as a sanity check: whether the qualitative representation patterns reported in that literature reoccur in hard-target C51 and target-free PQN.

Across all 26 games with 3 seeds each at UTD 4, unnormalized C51 has the highest dormant fraction and lowest effective rank~(\Cref{fig:combined-plasticity}). Both BN variants have lower dormant-fraction IQMs than their LayerNorm counterparts throughout training and higher effective-rank IQMs after the first diagnostic. The two WN variants remain close to their unprojected counterparts relative to the difference between normalizers.

The PQN diagnostics continue to evolve over 400M frames. Within each matched loss, BN generally has a lower dormant-fraction IQM than LayerNorm, and the HL-Gauss BN variant has a higher effective-rank IQM than its LayerNorm counterpart. Scalar BN and scalar LayerNorm have more similar effective ranks.
The shared ordering reproduces the broad qualitative pattern reported in prior normalization studies~\citep{palenicek2025crossqwn,palenicek2026xqc}, but remains a descriptive regularity rather than a mechanism for the return results.

\section{Conclusion}
\label{sec:conclusion}

In this paper, we demonstrate that Batch normalization can substantially improve discrete-action value learning and is not uniformly harmful in value learning as often reported.
Its success depends on the forward mode and careful application.
If applied correctly, however, it can provide similar sample-efficiency benefits as previously reported in actor-critic settings.
In C51, replacing stored target moments with current-minibatch statistics reverses BN degradation and allows stable update-to-data ratio scaling up to 12.
In PQN, changing the acting and bootstrap protocols recovers BN performance even without replay or a separate target network. Across 26 Atari games at 400M frames, the selected batch-statistic configuration achieves comparable aggregate performance over training and a higher final aggregate score than LayerNorm.
Together, these findings challenge the prevailing view that BN is ill-suited to discrete-action value learning and show that its reported failures can be reversed through appropriate forward-mode choices.

\bibliographystyle{assets/plainnat}
\bibliography{paper}

\appendix

\section{Implementation Details}
\label{app:implementation}

This section specifies evaluation, Atari preprocessing, network architectures, optimizers, and training schedules that complement the role-specific BatchNorm semantics above.

\subsection{Evaluation and uncertainty}
\label{sec:evaluation}

We transform returns to human-normalized scores,
\begin{equation}
\operatorname{HNS}(g)=
\frac{R(g)-R_{\mathrm{random}}(g)}
     {R_{\mathrm{human}}(g)-R_{\mathrm{random}}(g)},
\label{eq:hns}
\end{equation}
and aggregate them with the interquartile mean (IQM)~\citep{mnih2015dqn,agarwal2021rliable}. C51 AUC is the arithmetic mean HNS over each run's evaluation grid, and the last evaluation occurs at 5M frames. Marginal 95\% intervals use 5,000 seed-stratified resamples conditional on the fixed 26-game set, while target-mode contrasts resample matched seeds within game. Relative UTD results divide by unnormalized C51 at UTD $0.25$.

The Pong diagnostic uses 30 frame bins; AUC averages all bins, the final statistic averages the last 4M frames, and error bars are standard errors over five seeds. The 400M PQN archives contain 60 complete bins for every one of the ten seeds in each displayed arm--game cell. AUC averages the 60 human-normalized bin values within each run, while the endpoint uses the last bin. AUC, final IQM, curve intervals, and paired differences use 5,000 within-game seed resamples; paired contrasts use the same sampled seed identities for both arms.

\subsection{Atari environments}

Both studies use the same 26 games: Alien, Amidar, Assault, Asterix, BankHeist, BattleZone, Boxing, Breakout, ChopperCommand, CrazyClimber, DemonAttack, Freeway, Frostbite, Gopher, Hero, Jamesbond, Kangaroo, Krull, KungFuMaster, MsPacman, Pong, PrivateEye, Qbert, RoadRunner, Seaquest, and UpNDown. Human normalization uses the random and human reference scores of \citet{mnih2015dqn}, following \citet{agarwal2021rliable}.

\begin{figure}[t!]
    \centering
    \includegraphics[width=\textwidth]{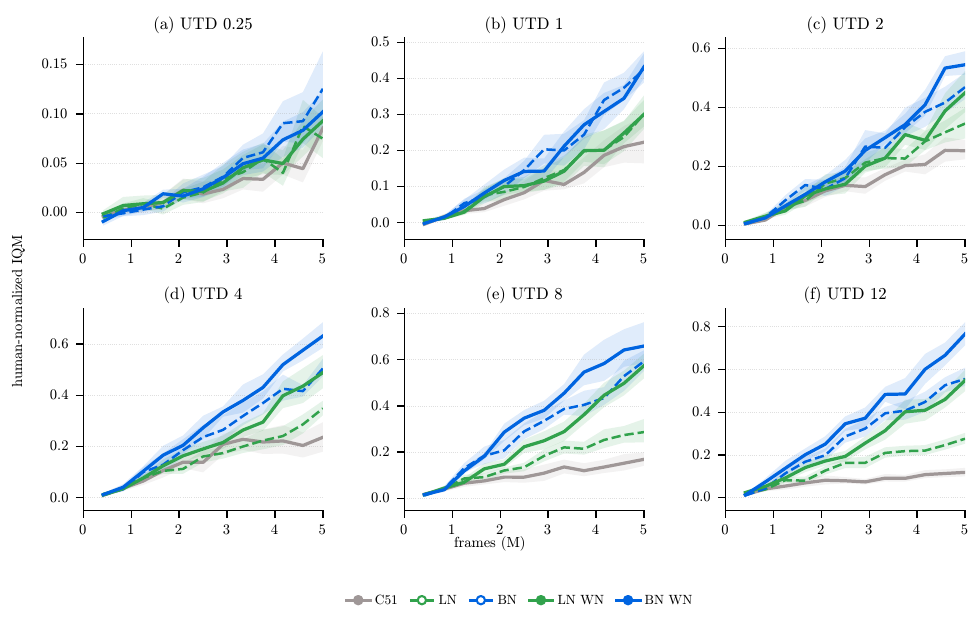}
    \caption{\textbf{Aggregate C51 learning curves across replay ratios.}
    Human-normalized IQM over 26 games and three seeds at UTD $0.25$, $1$, $2$, $4$, $8$, and $12$. Each panel uses its own vertical scale. All BN arms use batch statistics at the bootstrap forward; ``WN'' denotes the fixed-norm projection. C51, LayerNorm, and BN use a constant learning rate, while the two fixed-norm-projection arms (LN WN, BN WN) use a linear-decay learning rate, matching \Cref{fig:c51-main}. Bands are marginal 95\% seed-stratified bootstrap intervals conditional on the fixed game set. These curves are the longitudinal counterparts of \Cref{fig:c51-main}c,d.}
    \label{fig:c51-utd-curves}
\end{figure}

\FloatBarrier
\clearpage

\subsection{Hyperparameter tables}
\label{app:hyperparams}

For reference, \Cref{tab:c51-hparams} 
collects the C51 configuration and \Cref{tab:pqn-hparams} 
the PQN configuration. Values are as run; where the code retains a CleanRL default that the launchers override (e.g.\ the total budget and warm-up), the as-run value is listed.

\begin{table}[t]
\centering
\footnotesize
\caption{C51 shared training hyperparameters (Atari 26-game suite).}
\label{tab:c51-hparams}
\begin{tabular}{@{}ll@{}}
\toprule
Hyperparameter & Value \\
\midrule
\multicolumn{2}{@{}l}{\emph{Architecture}}\\
Convolutional encoder & Conv$(32,8{\times}8,\mathrm{s}4)$--Conv$(64,4{\times}4,\mathrm{s}2)$--Conv$(64,3{\times}3,\mathrm{s}1)$ \\
Dense (penultimate) layer & 512 units \\
Output head & per-action, 51 atoms \\
Hidden activation & ReLU \\
Initialization & Kaiming-normal (ReLU gain); biases $0$ \\
\midrule
\multicolumn{2}{@{}l}{\emph{Distributional objective}}\\
Atoms & 51 \\
Support $[v_{\min},v_{\max}]$ & $[-10,10]$ \\
Bootstrap & one-step return \\
\midrule
\multicolumn{2}{@{}l}{\emph{Optimizer}}\\
Optimizer & Adam \\
Learning rate & $2.5\times10^{-4}$ \\
Adam $\epsilon$ & $0.01/32\approx3.13\times10^{-4}$ \\
Adam $(\beta_1,\beta_2)$ & $(0.9,0.999)$ \\
Discount $\gamma$ & $0.99$ \\
\midrule
\multicolumn{2}{@{}l}{\emph{Replay and updates}}\\
Replay buffer size & $10^{6}$ transitions \\
Minibatch size & $32$ \\
Warm-up (learning starts) & $20{,}000$ decisions \\
Target network update & hard copy every $10{,}000$ decisions \\
Total budget & $1.25$M decisions ($=5$M frames) \\
Parallel environments & $1$ \\
\midrule
\multicolumn{2}{@{}l}{\emph{Exploration ($\epsilon$-greedy)}}\\
Start\,$\to$\,end $\epsilon$ & $1.0\to0.01$ \\
Decay fraction & first $10\%$ of training \\
\midrule
\multicolumn{2}{@{}l}{\emph{Evaluation}}\\
Frequency & every $100{,}000$ decisions \\
Episodes per evaluation & $10$ \\
Evaluation $\epsilon$ & $0.001$ \\
Seeds per game & $3$ (over $26$ games) \\
\bottomrule
\end{tabular}
\end{table}

\begin{table}[t]
\centering
\footnotesize
\caption{PQN hyperparameters (target-free, on-policy; $40$M Pong diagnostic and $400$M Atari-26 experiment).}
\label{tab:pqn-hparams}
\begin{tabular}{@{}ll@{}}
\toprule
Hyperparameter & Value \\
\midrule
\multicolumn{2}{@{}l}{\emph{Architecture}}\\
Convolutional encoder & Conv$(32,8,4)$--Conv$(64,4,2)$--Conv$(64,3,1)$ \\
Dense (penultimate) layer & 512 units \\
Normalization & every hidden layer (LayerNorm or BatchNorm); head unnormalized \\
Output head & scalar (regression) or 51-atom (HL-Gauss) \\
\midrule
\multicolumn{2}{@{}l}{\emph{Optimizer}}\\
Optimizer & RAdam \\
Learning rate & $2.5\times10^{-4}$ (constant) \\
Gradient clipping & global norm $10$ \\
Discount $\gamma$ & $0.99$ \\
$Q(\lambda)$ trace $\lambda$ & $0.65$ \\
\midrule
\multicolumn{2}{@{}l}{\emph{Rollout and updates (no replay, no target network)}}\\
Parallel environments & $128$ \\
Rollout length & $32$ steps \\
Transitions per update & $128\times32=4096$ \\
Minibatches per epoch & $32$ (size $128$) \\
Epochs per update & $2$ ($=64$ optimizer steps/update) \\
\midrule
\multicolumn{2}{@{}l}{\emph{Exploration ($\epsilon$-greedy)}}\\
Start\,$\to$\,end $\epsilon$ & $1.0\to0.001$ \\
Decay fraction & first $10\%$ of updates \\
\midrule
\multicolumn{2}{@{}l}{\emph{BatchNorm}}\\
BN $\epsilon$ & $10^{-5}$ \\
Running momentum (Flax) & $0.99$ (running baseline); batch arms $0.9$ \\
\midrule
\multicolumn{2}{@{}l}{\emph{HL-Gauss objective (matched-objective arms)}}\\
Bins & $51$ \\
Target clip & $[-10,20]$ \\
Gaussian width & $0.75$ bins \\
\midrule
\multicolumn{2}{@{}l}{\emph{Horizons and evaluation}}\\
Pong diagnostic & $40$M frames ($=10$M transitions), $5$ seeds \\
Atari-26 experiment & $400$M frames ($=100$M transitions), $10$ seeds \\
Coupled greedy test & $8$ additional zero-$\epsilon$ environments \\
\bottomrule
\end{tabular}
\end{table}

\FloatBarrier
\clearpage

\section{Additional C51 Results}
\label{app:c51-secondary}

\subsection{Per-environment learning curves}

The following grids report the complete per-game view behind the aggregate UTD sweep. Each grid uses the same three seed identities and 5M-frame horizon as the corresponding point in \Cref{fig:c51-main}c,d.

\begin{figure}[h!]
    \centering
    \includegraphics[width=0.93\textwidth]{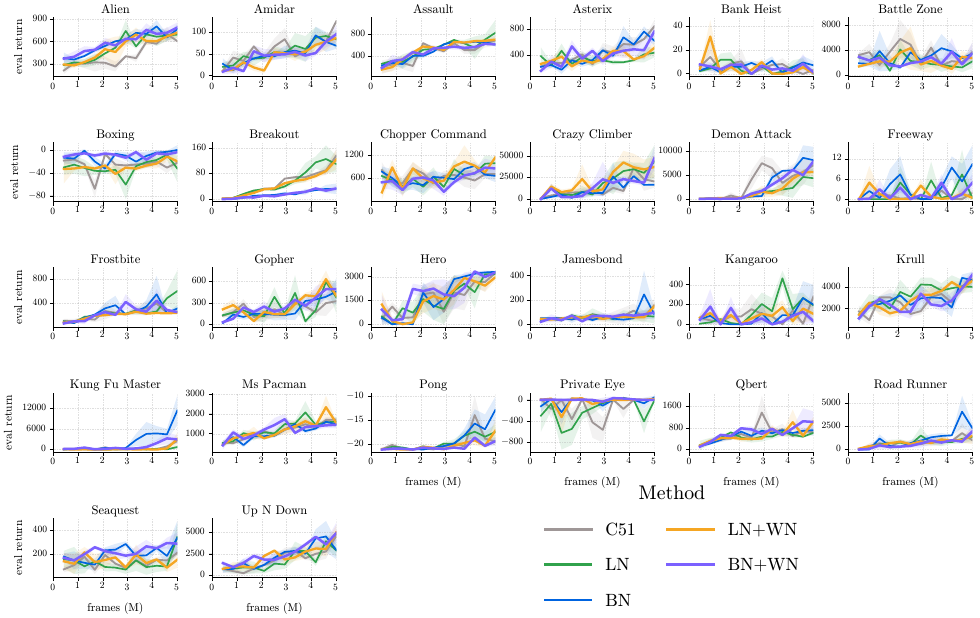}
    \caption{\textbf{Per-environment C51 learning curves at UTD $0.25$.}
    Curves are raw evaluation returns, averaged over three seeds; bands are one standard error. All BN arms use batch statistics at the bootstrap forward. WN denotes the fixed-norm projection. C51, LayerNorm, and BN use a constant learning rate, while LN+WN and BN+WN use a linear-decay learning rate, matching \Cref{fig:c51-main}.}
    \label{fig:c51-pergame-025}
\end{figure}

\clearpage

\begin{figure}[t!]
    \centering
    \includegraphics[width=0.93\textwidth]{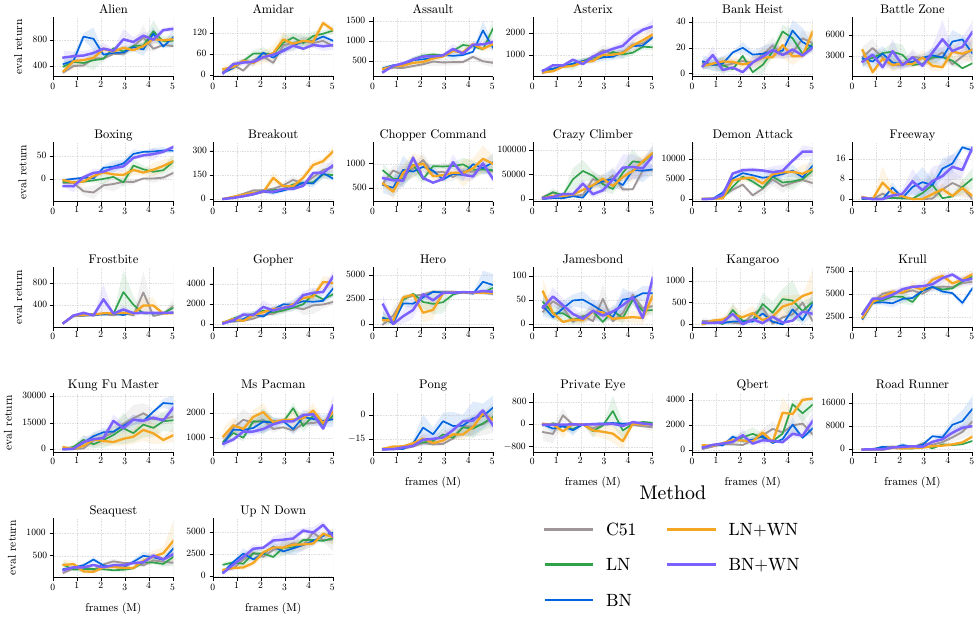}
    \par\vspace{4pt}
    \includegraphics[width=0.93\textwidth]{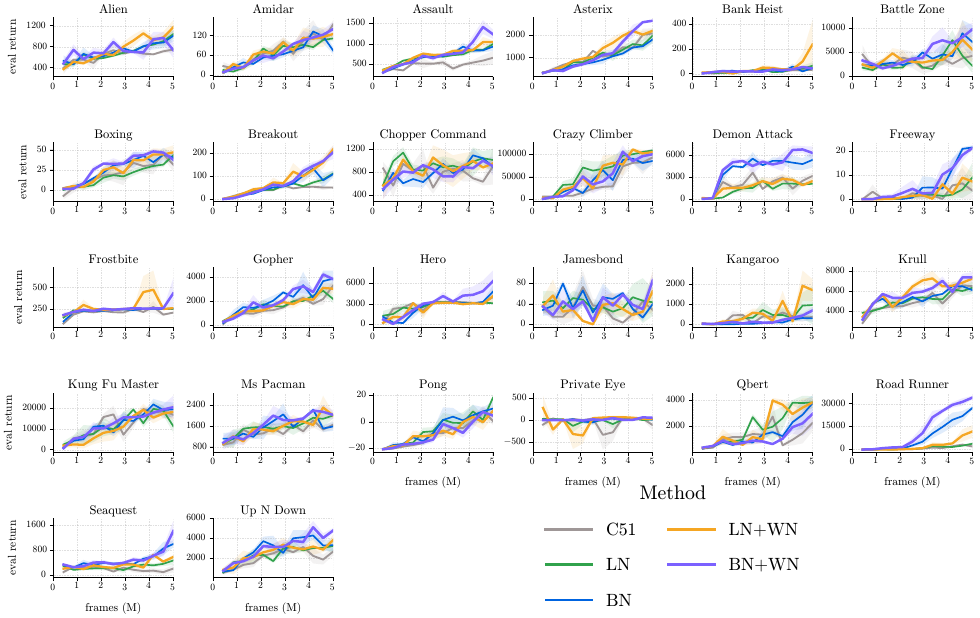}
    \caption{\textbf{Per-environment C51 learning curves at UTD $1$ and $2$.}
    The top grid reports UTD $1$ and the bottom grid UTD $2$. Curves are raw evaluation returns, averaged over three seeds; bands are one standard error. Conventions follow \Cref{fig:c51-pergame-025}.}
    \label{fig:c51-pergame-1-2}
\end{figure}

\clearpage

\begin{figure}[t!]
    \centering
    \includegraphics[width=0.93\textwidth]{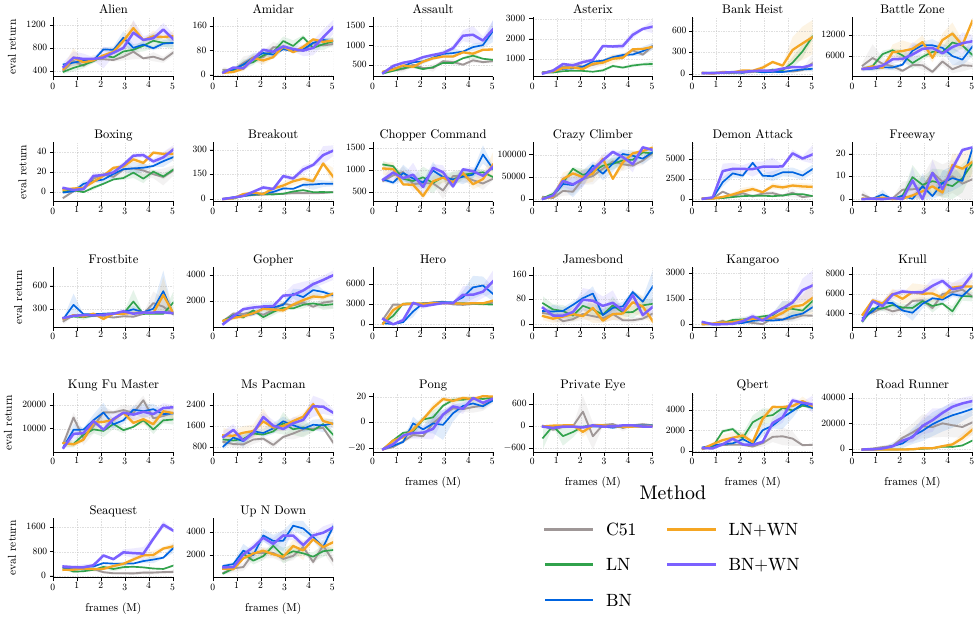}
    \par\vspace{4pt}
    \includegraphics[width=0.93\textwidth]{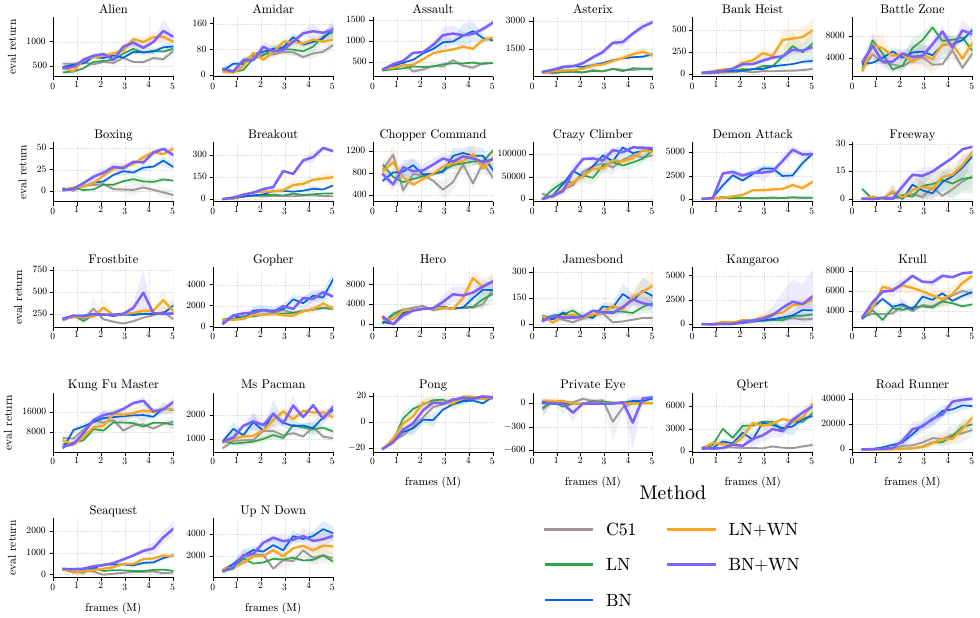}
    \caption{\textbf{Per-environment C51 learning curves at UTD $4$ and $8$.}
    The top grid reports UTD $4$ and the bottom grid UTD $8$. Curves are raw evaluation returns, averaged over three seeds; bands are one standard error. Conventions follow \Cref{fig:c51-pergame-025}.}
    \label{fig:c51-pergame-4-8}
\end{figure}

\clearpage

\begin{figure}[t!]
    \centering
    \includegraphics[width=0.93\textwidth]{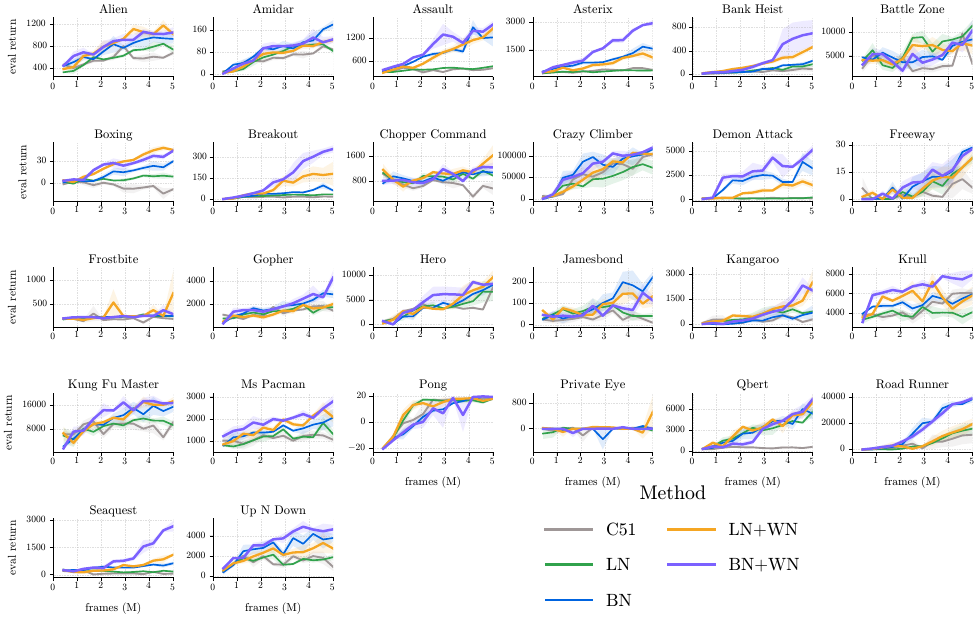}
    \caption{\textbf{Per-environment C51 learning curves at UTD $12$.}
    Curves are raw evaluation returns, averaged over three seeds; bands are one standard error. Conventions follow \Cref{fig:c51-pergame-025}.}
    \label{fig:c51-pergame-12}
\end{figure}

\clearpage

\section{Additional PQN Results}
\label{app:pqn-additional}

\begin{figure}[h!]
    \centering
    \includegraphics[width=\textwidth]{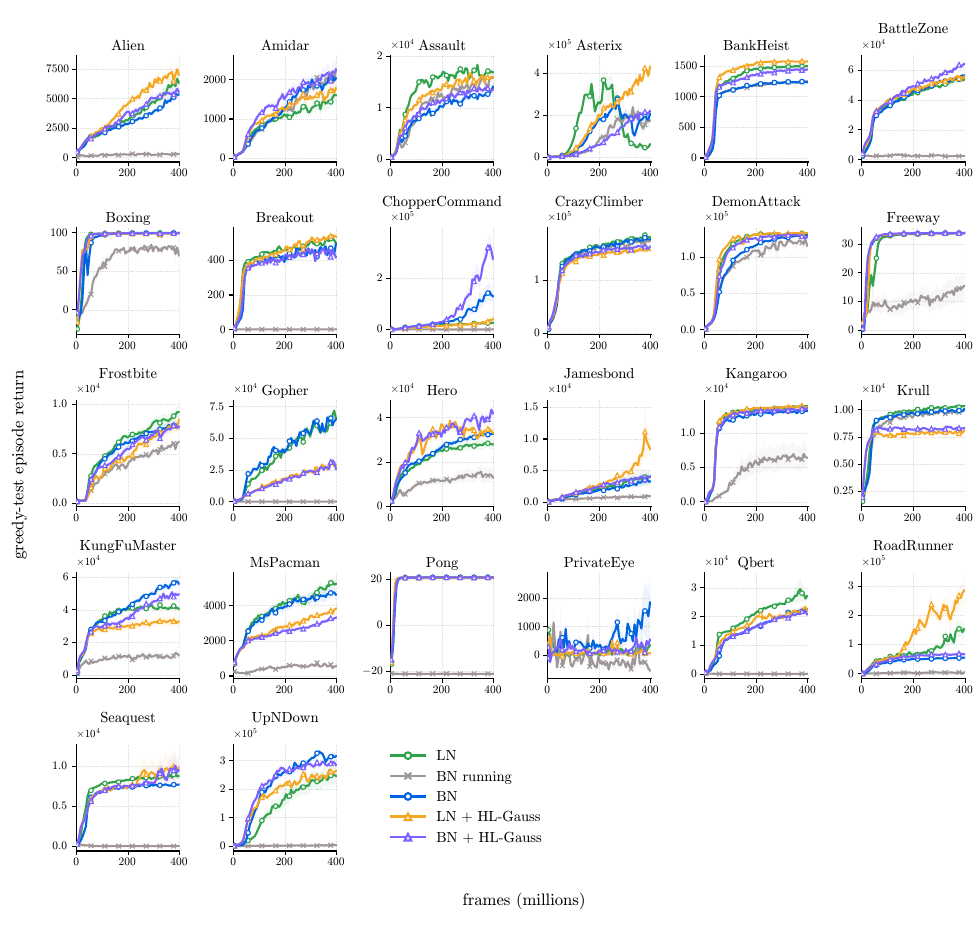}
    \caption{\textbf{Per-environment coupled greedy-test learning curves over 400M Atari frames.}
    For each game, curves show the mean raw return and one standard error across all ten seeds.
    The gray curve (crosses) shows the released running-statistic BN configuration from \citet{gallici2025simplifying}.
    Green and orange identify scalar-regression and HL-Gauss LayerNorm; blue and purple show the corresponding batch-statistic BN configurations. Circles and triangles distinguish the scalar and HL-Gauss objectives.}
    \label{fig:pqn-pergame-balanced}
\end{figure}

\Cref{fig:pqn-pergame-balanced} shows the PQN per-game learning curves.

\FloatBarrier

\clearpage

\section{Discrete-Action Actor-Critic Instantiation on Atari}
\label{app:mrq}

The main paper studies BatchNorm forward modes in \emph{value-based} learners. For completeness, we report an actor--critic instantiation of the same BN recipe for discrete actions using MR.Q's~\citep{fujimoto2025mrq} Gumbel-Softmax trick.
XQC~\citep{palenicek2026xqc} pairs a CrossQ-style batch-statistic BN critic with a fixed-norm weight projection in a TD3-style actor--critic agent; we run it on discrete-action Atari against two baselines: MR.Q~\citep{fujimoto2025mrq}, a stronger and considerably more complex agent, and TD3~\citep{fujimoto2018td3}. All three use the same protocol: 26 games, five seeds, and $5\times10^{6}$ frames ($1.25\times10^{6}$ policy steps at action repeat 4), with evaluation every $2\times10^{5}$ frames. These results lie outside the paper's value-based scope and are included only to indicate that the recipe transfers to the actor--critic setting; they are not part of the paper's claims.

\begin{figure}[h]
    \centering
    \includegraphics[width=0.62\textwidth]{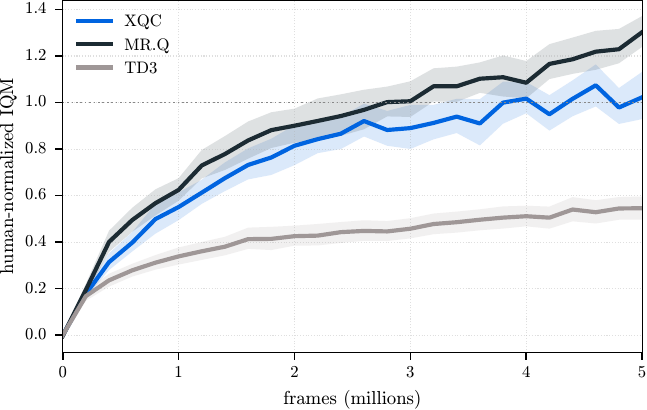}
    \caption{\textbf{Actor--critic Atari aggregate.} Human-normalized IQM over 26 games (five seeds) for XQC, MR.Q, and TD3 to 5M frames ($5\times10^{6}$ frames at action repeat 4). Bands are 95\% seed-bootstrap intervals; the dotted line marks human-level performance.}
    \label{fig:mrq-agg}
\end{figure}

\begin{figure}[h]
    \centering
    \includegraphics[width=\textwidth]{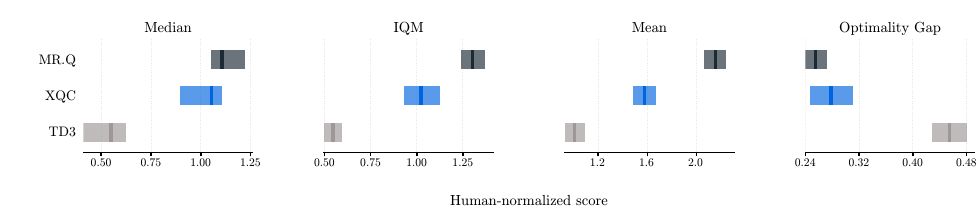}
    \caption{\textbf{Aggregate metrics (actor--critic Atari).} Median, IQM, mean, and optimality gap of the human-normalized score at 5M frames over 26 games (five seeds), with 95\% stratified bootstrap intervals~\citep{agarwal2021rliable}. Higher is better except for the optimality gap.}
    \label{fig:mrq-agarwal}
\end{figure}

\begin{figure}[h]
    \centering
    \includegraphics[width=\textwidth]{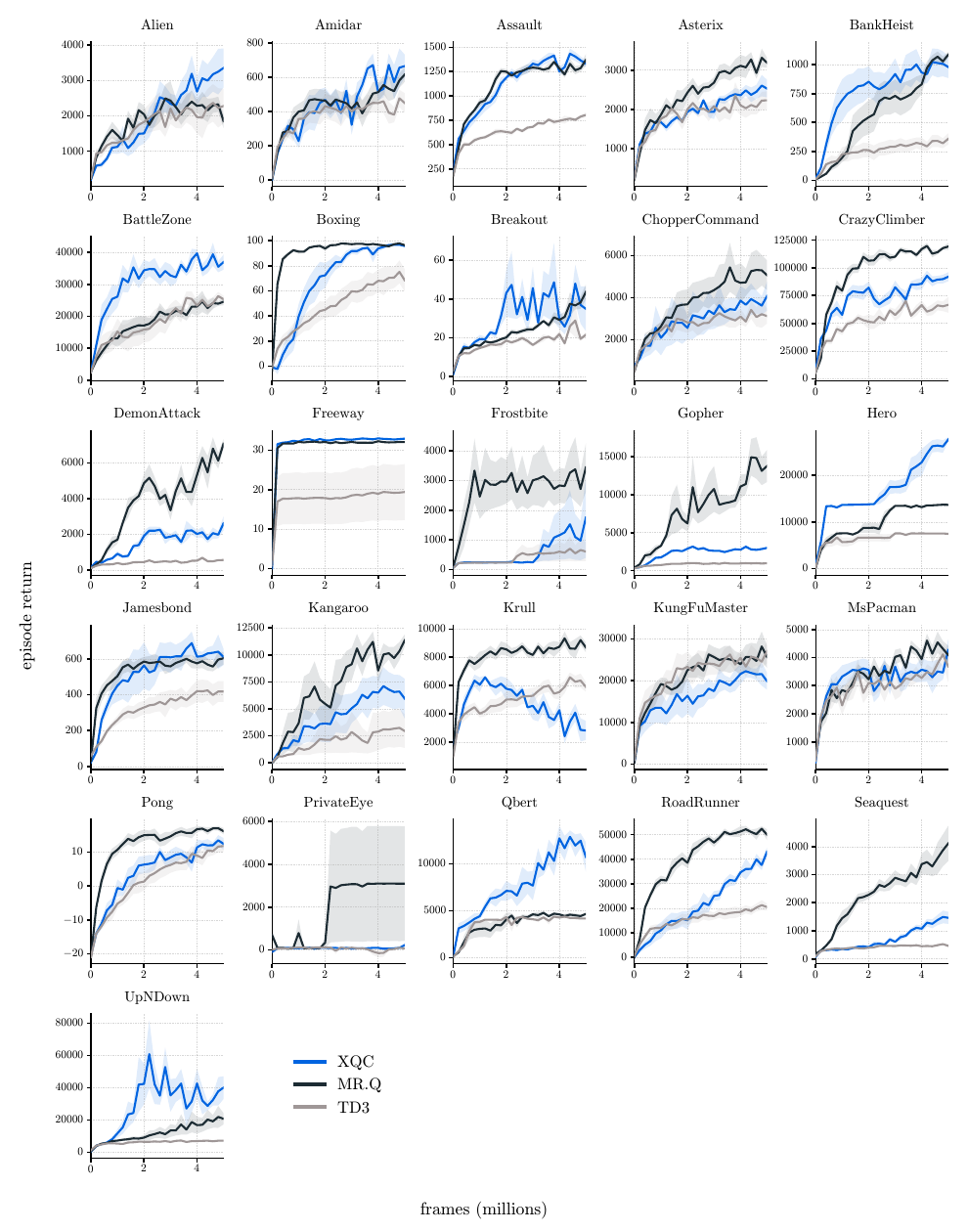}
    \caption{\textbf{Per-environment actor--critic Atari curves.} Raw evaluation return for XQC, MR.Q, and TD3 on each of the 26 games, averaged over five seeds; bands are one standard error. Evaluation is every $2\times10^{5}$ frames to $5\times10^{6}$ frames.}
    \label{fig:mrq-pergame}
\end{figure}

\Cref{fig:mrq-agg} reports the human-normalized IQM over the 26 games. At 5M frames the final IQMs are $1.30$ (95\% interval $[1.24,1.37]$) for MR.Q, $1.02$ ($[0.93,1.13]$) for XQC, and $0.55$ ($[0.50,0.60]$) for TD3: XQC reaches roughly human-level aggregate performance and clearly exceeds TD3, while the more complex MR.Q remains ahead. \Cref{fig:mrq-agarwal} reports four aggregate metrics with 95\% stratified bootstrap intervals~\citep{agarwal2021rliable}: XQC and MR.Q are close on the median ($1.05$ versus $1.11$) and optimality gap ($0.28$ versus $0.25$), while MR.Q's advantage is larger on the IQM and mean, which weight its large gains on high-ceiling games. The per-environment curves in \Cref{fig:mrq-pergame} show the heterogeneity behind the aggregate---XQC exceeds TD3 on 24 of the 26 games and outperforms MR.Q on 9.

\FloatBarrier

\end{document}